\documentclass[lettersize,journal]{IEEEtran}
\usepackage{amsmath,amsfonts}
\usepackage{algorithmic}
\usepackage{algorithm}
\usepackage{array}
\usepackage{textcomp}
\usepackage{stfloats}
\usepackage{url}
\usepackage{verbatim}
\usepackage{graphicx}
\usepackage{cite}

\usepackage{orcidlink}
\usepackage{multirow}%
\usepackage{booktabs}%
\usepackage{pifont}
\usepackage{subcaption}

\newcommand{\cmark}{\ding{51}}%

\begin{document}

\title{Evaluating Facial Expression Recognition Datasets for Deep Learning: A Benchmark Study with Novel Similarity Metrics}



\author{
    F. Xavier Gaya-Morey \orcidlink{0000-0003-1231-7235}, 
    Cristina Manresa-Yee \orcidlink{0000-0002-8482-7552}, 
    Célia Martinie \orcidlink{0000-0001-7907-3170}, 
    Jose M. Buades-Rubio \orcidlink{0000-0002-6137-9558}

    
    \thanks{F. Xavier Gaya-Morey, Cristina Manresa-Yee and Jose M. Buades-Rubio are with the Computer Graphics and Vision and AI Group (UGIVIA), at the Universitat de les Illes Balears, Carretera de Valldemossa, km 7.5, Palma, 07122, Illes Balears, Spain (e-mail: francesc-xavier.gaya@uib.es; cristina.manresa@uib.es; josemaria.buades@uib.es).}
    
    \thanks{Célia Martinie is with the ICS-IRIT, at the University Toulouse 3, Paul Sabatier, 118 Rte de Narbonne, Toulouse, 31062, France (e-mail: celia.martinie@irit.fr).}
}




\maketitle

\begin{abstract}
This study investigates the key characteristics and suitability of widely used Facial Expression Recognition (FER) datasets for training deep learning models. In the field of affective computing, FER is essential for interpreting human emotions, yet the performance of FER systems is highly contingent on the quality and diversity of the underlying datasets. To address this issue, we compiled and analyzed 24 FER datasets--including those targeting specific age groups such as children, adults, and the elderly--and processed them through a comprehensive normalization pipeline. In addition, we enriched the datasets with automatic annotations for age and gender, enabling a more nuanced evaluation of their demographic properties. To further assess dataset efficacy, we introduce three novel metrics--Local, Global, and Paired Similarity--which quantitatively measure dataset difficulty, generalization capability, and cross-dataset transferability. Benchmark experiments using state-of-the-art neural networks reveal that large-scale, automatically collected datasets (e.g., AffectNet, FER2013) tend to generalize better, despite issues with labeling noise and demographic biases, whereas controlled datasets offer higher annotation quality but limited variability. Our findings provide actionable recommendations for dataset selection and design, advancing the development of more robust, fair, and effective FER systems.
\end{abstract}


\section{Introduction}

    Facial Expression Recognition (FER) is a key research area within affective computing, focused on identifying and interpreting human emotions through facial cues. While facial expressions themselves do not equate to emotions~\cite{barrett2019emotional}, they serve as vital visual indicators of emotional states, facilitating non-verbal communication and enhancing social interactions~\cite{degelder}. Although the universality of facial expressions across cultures remains a topic of debate~\cite{Barrett}, Ekman’s foundational work on six fundamental expressions--anger, happiness, surprise, disgust, sadness, and fear--continues to be a cornerstone in this field~\cite{ekmanuniversal}. The increasing relevance of FER is reflected in its projected market growth, estimated to reach \$682.2 billion by 2032, driven by applications in medical diagnostics~\cite{Grabowski2019}, human behavior analysis~\cite{Barreto2017}, and human-computer interaction~\cite{Ramis2020}.
    
    Traditional machine learning methods for facial expression recognition relied on Action Units (AUs) and handcrafted feature descriptors~\cite{mohana2024facial}. AUs represent facial muscle activations and were used by human observers in the Facial Action Coding System (FACS)~\cite{ekman1978manual} to describe emotional expressions with high accuracy. However, with the arrival of Deep Learning (DL) methods--capable of automatically extracting feature representations from raw images and mimicking human visual perception--these approaches have become the standard in FER due to their superior accuracy and generalization capabilities~\cite{mohana2024facial, li2022deep}. Given the data-intensive nature of DL, numerous FER datasets have been developed. However, the performance of trained models is highly dependent on the dataset used for training, making dataset selection a critical factor. Characteristics such as dataset size, recording conditions, labeling methodology, and participant demographics must be carefully considered to mitigate potential biases. For instance, significant differences in facial expression recognition exist between age groups, such as children, adults, and the elderly~\cite{fölster2014facial}. Despite this, large-scale datasets like AffectNet~\cite{mollahosseini2017affectnet} do not provide age distribution information, and most existing datasets focus on specific age groups. For example, LIRIS-CSE~\cite{khan2019novel} targets children, KDEF~\cite{d_1998karolinska} focuses on adults, and ElderReact~\cite{ma2019elderreact} is designed for elderly individuals.

    To address these gaps, we have compiled and analyzed 24 FER datasets, including those tailored to specific age groups. These datasets were normalized through multiple processing steps, and we enriched them with additional annotations--such as estimated age and gender--using deep learning models. This allowed for an in-depth examination of current FER dataset characteristics. Additionally, we introduced three novel metrics to assess different aspects of the datasets and conducted a benchmark by training state-of-the-art deep learning models on each dataset. To our knowledge, this study constitutes the most comprehensive compendium of FER datasets to date, providing an extensive evaluation that will be valuable for future research. For transparency and reproducibility, we have made publicly available the list of images used from each database, the computed automatic annotations, and the code\footnote{\url{https://github.com/Xavi3398/fer_benchmark}}.

    The remainder of this paper is structured as follows. First, we review previous FER benchmarks and commonly used datasets. We then outline the primary and secondary research questions guiding our study. Next, we describe our methodology, including dataset collection and normalization, metric definitions, and benchmark settings. The results are subsequently presented and analyzed, grouped according to the research questions. Finally, we discuss key findings, provide recommendations for dataset selection and construction, and conclude the study.

\section{Related Work}

    \subsection{Common Facial Expression Recognition Datasets}

        Facial expression recognition has been a focus of research for many years, not only for developing automated methods but also for psychological experiments, leading to the availability of numerous datasets. In our literature review, we identified 28 common datasets within the field, with particular attention to those including older adults and children--two underrepresented age groups due to their higher vulnerability. These datasets, which serve both automated FER research and controlled psychological experiments, vary widely in terms of collection methods, demographics, and expression types. 
        
        In Table~\ref{tab:datasets}, we summarize the key characteristics of the datasets, published between 1997 and 2022. This compilation of datasets illustrates the diversity and evolution of FER resources, ranging from controlled laboratory collections to large-scale, ``in the wild" data. The varied characteristics--including subject demographics, expression types, collection methods, and imaging modalities--underscore the importance of dataset selection and curation in developing robust and generalizable facial expression recognition systems.
        
        \begin{table}[h]
        \caption{Description of the facial expression recognition datasets found}
        \label{tab:datasets}
        \centering
        \setlength{\tabcolsep}{3pt}
        \resizebox{\columnwidth}{!}{
        \begin{tabular}{llc|rccr|ccccccc}
            \toprule
            \multicolumn{3}{c|}{\textbf{Details}} & \multicolumn{4}{c|}{\textbf{Data}} & \multicolumn{7}{c}{\textbf{Classes}} \\
            \textbf{Ref.} & \textbf{Name} & \textbf{Year} & \textbf{Users} & \textbf{Ages} & \textbf{Type}  & \textbf{Samples} & \multicolumn{1}{c}{\rotatebox{90}{\textbf{Anger}}} & \multicolumn{1}{c}{\rotatebox{90}{\textbf{Disgust}}} & \multicolumn{1}{c}{\rotatebox{90}{\textbf{Fear}}} & \multicolumn{1}{c}{\rotatebox{90}{\textbf{Happiness}}} & \multicolumn{1}{c}{\rotatebox{90}{\textbf{Sadness}}} & \multicolumn{1}{c}{\rotatebox{90}{\textbf{Surprise}}} & \multicolumn{1}{c}{\rotatebox{90}{\textbf{Neutral}}} \\ 
            \midrule
            \cite{ramis2022novel}             & FEGA                  & 2022 & 51    & 21-66     & Image & 2,856   & \cmark & \cmark  & \cmark & \cmark    & \cmark  & \cmark   & \cmark  \\
            \cite{ramis2022novel}             & FE-Test               & 2022 & N/A   & N/A       & Image & 210     & \cmark & \cmark  & \cmark & \cmark    & \cmark  & \cmark   & \cmark  \\
            \cite{nfhi2020}                   & NHFI                  & 2020 & N/A   & N/A       & Image & 5,558   & \cmark & \cmark  & \cmark & \cmark    & \cmark  & \cmark   & \cmark  \\
            \cite{yang2020tsinghua}           & Tsinghua              & 2020 & 110   & 18-76     & Image & 1,128   & \cmark & \cmark  & \cmark & \cmark    & \cmark  & \cmark   & \cmark  \\
            \cite{ma2019elderreact}           & ElderReact            & 2019 & 46    & N/A       & Video & 1,323   & \cmark & \cmark  & \cmark & \cmark    & \cmark  & \cmark   &         \\
            \cite{khan2019novel}              & LIRIS-CSE                 & 2019 & 12    & 4-12      & Video & 208     & \cmark & \cmark  & \cmark & \cmark    & \cmark  & \cmark   &         \\
            \cite{zhang2018facial}            & ExpW                  & 2018 & N/A   & N/A       & Image & 91,793  & \cmark & \cmark  & \cmark & \cmark    & \cmark  & \cmark   & \cmark  \\
            \cite{mollahosseini2017affectnet} & AffectNet             & 2017 & N/A   & N/A       & Image & 440,000 & \cmark & \cmark  & \cmark & \cmark    & \cmark  & \cmark   & \cmark  \\
            \cite{meuwissen2017creation}      & DEFSS                 & 2017 & 116   & 8-30      & Image & 404     & \cmark &         & \cmark & \cmark    & \cmark  &          & \cmark  \\
            \cite{li2017reliable}             & RAF-DB                & 2017 & N/A   & 0-70      & Image & 29,672  & \cmark & \cmark  & \cmark & \cmark    & \cmark  & \cmark   & \cmark  \\
            \cite{wang2016database}           & Aff. Int. & 2016 & 16    & N/A       & Video & 810     & \cmark & \cmark  &        & \cmark    & \cmark  &          & \cmark  \\
            \cite{zhang2016biovid}           & BioVidEmo            & 2016 & 90    & 18-65     & Video & 430     & \cmark & \cmark  & \cmark & \cmark    & \cmark  &          &         \\
            \cite{nojavanasghari2016emoreact} & EmoReact              & 2016 & 63    & 4-14      & Video & 1,102   &        & \cmark  & \cmark & \cmark    &         & \cmark   &         \\
            \cite{lobue2015child}             & CAFE                  & 2015 & 154   & 2-8       & Image & 1,192   & \cmark & \cmark  & \cmark & \cmark    & \cmark  & \cmark   & \cmark  \\
            \cite{olszanowski2014warsaw}      & WSEFEP                & 2014 & 30    & 20-30     & Image & 210     & \cmark & \cmark  & \cmark & \cmark    & \cmark  & \cmark   & \cmark  \\
            \cite{dalrymple2013dartmouth}     & DDCF                  & 2013 & 80    & 6-16      & Image & 6,366   & \cmark & \cmark  & \cmark & \cmark    & \cmark  & \cmark   & \cmark  \\
            \cite{fer2013}                    & FER2013               & 2013 & N/A   & N/A       & Image & 35,887  & \cmark & \cmark  & \cmark & \cmark    & \cmark  & \cmark   & \cmark  \\
            \cite{dhall2007collecting}        & AFEW                  & 2012 & 330   & 1-70      & Video & 1,426   & \cmark & \cmark  & \cmark & \cmark    & \cmark  & \cmark   & \cmark  \\
            \cite{egger2011nimh}              & NIMH-ChEFS            & 2011 & 59    & 10-17     & Image & 482     & \cmark &         & \cmark & \cmark    & \cmark  &          & \cmark  \\
            \cite{dhall2011static}            & SFEW                  & 2011 & 95    & 1-70      & Image & 700     & \cmark & \cmark  & \cmark & \cmark    & \cmark  & \cmark   & \cmark  \\
            \cite{lucey2010extended}          & CK+                   & 2010 & 123   & 18-50     & Video & 593     & \cmark & \cmark  & \cmark & \cmark    & \cmark  & \cmark   & \cmark  \\
            \cite{ebner2010faces}             & FACES                 & 2010 & 171   & 19-80     & Image & 2,052   & \cmark & \cmark  & \cmark & \cmark    & \cmark  & \cmark   &         \\
            \cite{langner2010presentation}    & RaFD                  & 2010 & 67    & N/A       & Image & 8,040   & \cmark & \cmark  & \cmark & \cmark    & \cmark  & \cmark   & \cmark  \\
            \cite{yin2008high}                & BU-4DFE               & 2008 & 101   & 18-45     & Video & 606     & \cmark & \cmark  & \cmark & \cmark    & \cmark  & \cmark   &         \\
            \cite{pantic2005web}              & MMI                   & 2005 & 75    & 19-62     & Video & 2,900   & \cmark & \cmark  & \cmark & \cmark    & \cmark  & \cmark   &         \\
            \cite{minear2004lifespan}         & Lifespan              & 2004 & 576   & 18-93     & Image & 1,354   & \cmark & \cmark  &        & \cmark    & \cmark  & \cmark   & \cmark  \\
            \cite{d_1998karolinska}           & KDEF                  & 1998 & 70    & 20-30     & Image & 4,900   & \cmark & \cmark  & \cmark & \cmark    & \cmark  & \cmark   & \cmark \\
            \cite{lyons2020coding}            & JAFFE                 & 1997 & 10    & N/A       & Image & 219     & \cmark & \cmark  & \cmark & \cmark    & \cmark  & \cmark   & \cmark  \\
            \midrule
        \end{tabular}
        }
        \end{table}
        
    \subsection{Facial Expression Recognition Benchmarks}

        Despite the growing interest in facial expression recognition, there are few comprehensive benchmarks available in the literature. Some existing studies compare newly proposed datasets with previously established ones or conduct experiments across multiple datasets. For instance, Manresa et al.~\cite{ijimai} explored cross-dataset learning using five datasets--BU-4DFE, CK+, FEGA, JAFFE, and WSEFEP--demonstrating that generalization improves when training incorporates multiple datasets. Similarly, Chen et al.~\cite{chen2022cross-domain} conducted experiments across multiple datasets--CK+, JAFFE, SFEW, FER2013, ExpW, RAF-DB, and AFE--evaluating 14 algorithms to analyze and enhance cross-domain learning strategies.

        Several literature reviews have also compiled performance results across different datasets. For example, Li et al.~\cite{li2022deep} and Kopalidis et al.~\cite{kopalidis2024advances} present the performance of multiple methods on six and ten datasets, respectively. However, these approaches present several limitations. First, performance is often measured solely using accuracy, which is not a reliable metric for unbalanced datasets~\cite{branco2016survey}. Additionally, the number of expression classes varies across studies, meaning that the baseline performance of a random classifier is inconsistent. Furthermore, validation procedures differ between studies, employing methods such as leave-one-subject-out (LOSO), k-fold cross-validation with different values of \( k \), or pre-defined training-validation-test splits, which affects the performance estimation. Lastly, preprocessing techniques also vary, leading to inconsistencies in dataset evaluation. These factors underscore the need for a benchmark that normalizes datasets under the same conditions for a fair comparison.

        Many studies proposing new FER methods evaluate their approaches using multiple datasets. For instance, Dhall et al.~\cite{dhall2011static} introduced SFEW, a static version of AFEW, and validated their method on SFEW, JAFFE, and Multi-PIE~\cite{gross2010multi-pie}. Similarly, Wu et al.~\cite{wu2019facial} proposed a graph convolutional neural network and tested it on CK+ and JAFFE, while Li et al.~\cite{li2024fer-former} designed a FER-specific transformer architecture and evaluated it using RAF-DB, SFEW, and FER+~\cite{barsoum2016training}.
        
        Another relevant approach focuses on addressing misclassification errors in FER datasets. For example, Escobar et al.~\cite{mejia-escobar2023towards} tackled this issue by applying iterative training and automatic reclassification techniques on FER2013, NHFI, and AffectNet. Similarly, Liu et al.~\cite{liu2022uncertain} proposed auxiliary action unit graphs to mitigate misclassification errors, conducting experiments on RAF-DB and AffectNet.

\section{Research Questions}
\label{sec:rqs}

    In this section, we present the research questions addressed in this study, as outlined in Table~\ref{tab:questions}, and briefly explain the approach used to answer them. These questions are categorized into primary and secondary research questions.
    
    \begin{table}[h]
    \caption{Primary and secondary research questions}
    \label{tab:questions}
    \centering
    \resizebox{\columnwidth}{!}{
    \begin{tabular}{ll}
        \toprule
        \textbf{ID} & \textbf{Research Question} \\
        \midrule
        \multirow{2}{*}{\textbf{RQ1}} & \multirow{2}{25em}{\textbf{What are the main characteristics of commonly used FER datasets?}} \\[9pt]
        RQ1.1 & How were these datasets collected?\\
        RQ1.2 & What are their main demographic characteristics?\\
        \multirow{2}{*}{RQ1.3} & \multirow{2}{25em}{What are the most common expression classes, and how balanced are they?}\\[9pt]
        RQ1.4 & What are their key data characteristics?\\
        \midrule
        \multirow{2}{*}{\textbf{RQ2}} & \multirow{2}{25em}{\textbf{Which are the most suitable FER datasets for training deep learning models?}}\\[9pt]
        \multirow{2}{*}{RQ2.1} & \multirow{2}{25em}{How difficult are these datasets for deep learning models to learn from?}\\[9pt]
        \multirow{2}{*}{RQ2.2} & \multirow{2}{25em}{How well do models trained on these datasets generalize to others?}\\[9pt]
        \multirow{2}{*}{RQ2.3} & \multirow{2}{25em}{How redundant are these datasets when compared one-to-one?}\\[9pt]
        \bottomrule
    \end{tabular}%
    }
    \end{table}

    The first primary research question (RQ1) investigates the main characteristics of widely used FER datasets. Its associated secondary research questions explore specific aspects relevant to other studies, including the conditions under which the datasets were collected (e.g., sourced from the Internet, recorded with actors, etc.), their demographic characteristics (e.g., age, gender distributions), the labeled expression classes and their balance, and the key data characteristics, such as color information and whether the dataset consists of images or videos. 

    The second primary research question (RQ2) focuses on evaluating and comparing datasets for training deep learning models. The secondary research questions aim to assess three key aspects: the difficulty level of each dataset for deep learning models, the generalization capability of models trained on one dataset when tested on others, and the similarity between datasets in terms of model performance. 

\section{Methods}

    \subsection{Datasets Collection and Normalization}
    \label{sec:collection-normalization}

        We aimed to obtain as many datasets as listed in Table~\ref{tab:datasets} as possible. When datasets were publicly available, we accessed them via web forms, and for those without specified access procedures, we directly contacted the responsible individuals or the original authors via email. Despite our efforts, access to four datasets--Tsinghua, RAF-DB, Affective Interaction, and CAFE--was not granted, either due to a lack of response or denial of permission. Despite this, a public version of the RAF-DB dataset containing 15k images was found and used instead\footnote{\url{https://www.kaggle.com/datasets/shuvoalok/raf-db-dataset}}.

        The datasets presented several challenges due to inherent differences, such as varying numbers of classes, inconsistent data annotations (e.g., age, gender, head pose), different image resolutions, and differing face locations within images. To standardize and harmonize the data from these diverse sources, a six-step process was followed:

        \begin{enumerate}
            \item \textbf{Frame sampling}. Conversion of video datasets into image datasets by sampling frames.
            \item \textbf{Class unification}. Merging of similar expression labels.
            \item \textbf{Automatic labeling}. Generation of missing annotations for age, gender, and head pose.
            \item \textbf{Age group classification}. Categorization of each image into age groups, facilitating further experiments.
            \item \textbf{Exclusion of images}. Removal of low-quality and unannotated images.
            \item \textbf{Preprocessing}. Alignment of faces and normalization of image resolution, size, and color.
        \end{enumerate}

        Each of these steps is described in detail in the following sections.

        \subsubsection{Frame Sampling}

            Among the collected datasets, nine contain video data--ElderReact, EmoReact, BioVidEmo, AFEW, MMI, CK+, LIRIS-CSE, and BU-4DFE. Since this work focuses on facial expression recognition from still images, these datasets were sampled, extracting only a subset of frames, depending on the dataset type. For videos that transition between a neutral expression and one of the six basic emotions, three frames were extracted per video: one displaying the neutral expression and two depicting the target expression. Only a single neutral frame was taken from each video, as the neutral expression was consistently present, unlike the others. This sampling method was applied to the CK+, MMI, and BioVidEmo datasets. For videos in which the same expression is maintained throughout, five frames were sampled at equal intervals from the beginning to the end of each video. This approach was used for the ElderReact, EmoReact, and LIRIS-CSE datasets. For the AFEW dataset, we utilized its official static version, SFEW, eliminating the need for frame extraction. Similarly, the BU-4DFE dataset, as provided by the authors, was already pre-sampled and available in image format.

        \subsubsection{Class Unification}

            The datasets used in the benchmark included different classes, some of which referred to the same expression. The following classes were merged under the same label:

            \begin{itemize}
                \item \textbf{Anger:} ``arrabbiato'' (FEGA), ``annoyed'' (Lifespan), and ``grumpy'' (Lifespan)
                \item \textbf{Disgust:} ``disgusto'' (FEGA)
                \item \textbf{Fear:} ``afraid'' (DDCF, NIMH-ChEFS, KDEF), ``fearful'' (RaFD), and ``paura'' (FEGA)
                \item \textbf{Happiness:} ``joy'' (WSEFEP), ``allegria'' (FEGA), ``amusement'' and (BioVidEmo)
                \item \textbf{Sadness:} ``tristezza'' (FEGA)
                \item \textbf{Surprise:} ``sorpresa'' (FEGA)
                \item \textbf{Neutral:} ``neutra'' (FEGA), and ``profile'' (Lifespan)
            \end{itemize}
        
        \subsubsection{Automatic Labeling}
            \label{sec:automatic-labels}

            The available information across the datasets varied significantly, necessitating the use of multiple deep learning approaches to automatically annotate missing features such as age, gender, and head pose in the images.

            The MiVOLO transformer model~\cite{kuprashevich2024mivolo} was selected for age and gender estimation due to its demonstrated state-of-the-art performance. To obtain face crops, the default face detector from its official implementation, YOLOv8~\cite{jocher2023ultralytics}, was employed. Since age and gender are constant attributes for a single user or video clip (each containing only one individual), estimations were made for every image or frame associated with each user. The final age and gender labels were refined by taking the mode of the gender and the median of the age values across all images for each user, thus enhancing accuracy and resolving inconsistencies within the data.

            For head pose estimation, InsightFace~\cite{InsightFace} was utilized, employing ResNet50~\cite{resnet} for facial alignment. For consistency with the existing dataset annotations, images and frames were categorized into ``front", ``half\_left", ``half\_right", ``full\_left", ``full\_right" and ``back" classes. These categories correspond to approximate orientations of frontal (0 degrees), 45 degrees (left or right), 90 degrees (left or right), or over 90 degrees rotation of the head, respectively.

        \subsubsection{Age Group Extraction}

            We established three main age groups: 17 years or less (children), between 18 and 59 years (adults), and 60 years or more (elderly). We used the age annotations of the datasets when available to classify the images into one of these three groups. In many datasets, participants' age was not available, but the age group was, so we used it, matching it with the defined three age groups. In cases where neither age nor age group information was present in the dataset, the automatic age labels described in Section \ref{sec:automatic-labels} were used.
        
        \subsubsection{Image Exclusion}

            We excluded those images in which YOLOv8 failed to detect a face, as this step is essential for subsequent image preprocessing. Additionally, we introduced an exclusion criterion for non-frontal images. Specifically, images with head poses labeled as ``full\_left," ``full\_right," and ``back," as defined in Section \ref{sec:automatic-labels}, along with those lacking head pose annotations (indicating failure of automatic estimation), were removed. This step aimed to exclude challenging samples with poor camera perspectives that could negatively impact recognition performance.
    
        \subsubsection{Preprocessing}

            Before utilizing the images for training, a preprocessing step was applied to normalize the faces and facilitate the learning process of the models. Following the methodology outlined in previous studies~\cite{ijimai, gaya_morey2024unveiling, sabater_g_arriz2024pain, ramis2022novel}, facial landmarks were first detected using the estimator described in Section \ref{sec:automatic-labels} to obtain the coordinates of key facial points. The images were then rotated to horizontally align the eyes, using a line drawn between the coordinates of the eyes. Finally, the images were cropped, resized to a resolution of 224 × 224 pixels, and converted to grayscale to standardize the input format for the models.
    
    \subsection{Metrics Definition}
    \label{sec:metrics}

        In this study, we not only aimed to explore different datasets but also to assess their suitability for training deep learning models. To achieve this, we propose three new metrics that we have named Local Similarity, Global Similarity, and Paired Similarity, each capturing different aspects of the datasets based on a set of deep learning models. For simplicity, we use the following notations:
        
        \begin{itemize}
        
            \item Let \( M \) denote the set of all models, with \( |M| \) representing the number of models.
            
            \item Let \( D \) denote the set of all datasets, with \( |D| \) representing the number of datasets.
            
            \item Let $\hat{\theta}_{m, d_{train}, d_{test}}$ represent the estimation of the performance (F-1 score) of model \( m \in M \), trained on dataset \( d_{\text{train}} \in D \), and tested on dataset \( d_{\text{test}} \in D \).
            
            \item Given two datasets $d_1, d_2 \in D$ we define Cross Similarity function as $CS(d_1, d_2) = \frac{1}{|M|}\displaystyle \sum_{m \in M} {\hat{\theta}_{m, d_1, d_2}}$.
        \end{itemize}
        
        Local Similarity function $LS(d)$ quantifies the performance of models trained and tested on the same dataset $d$, thus providing a measure of how challenging it is for the models to learn the task from the dataset. It is computed using Equation~\ref{eq:difficulty}.
                
        \begin{equation}
            LS(d) = CS(d, d)
        \label{eq:difficulty}
        \end{equation}
        
        Global Similarity function $GS(d_{train})$ measures how well models trained on dataset $d_{train}$ perform when evaluated on all other datasets. It indicates the dataset’s ability to produce models with strong cross-dataset performance, i.e. it reflects how well a model trained on one dataset generalizes to others. The metric is computed using Equation~\ref{eq:generalization}.
            
        \begin{equation}
        \begin{split}
                    GS(d_{train}) = 
                    \frac{1}{|D|-1}{\displaystyle \sum_{
                    \substack{d_{test} \in D, \\ d_{test} \ne d_{train}}}{CS(d_{train}, d_{test})}}
        \end{split}
        \label{eq:generalization}
        \end{equation}
    
        Paired Similarity function $PS(d_{train}, d_{test})$ measures how well models trained on dataset $d_{train}$ perform on dataset $d_{test}$ relative to the difficulty of $d_{test}$ itself. This metric assesses the extent to which performance on $d_{test}$ can be achieved by training on $d_{train}$, providing insight into dataset redundancy. It is computed using Equation~\ref{eq:redundancy}.
            
        \begin{equation}
        \begin{split}
            PS(d_{train}, d_{test}) &=  \frac{CS(d_{train}, d_{test})}{CS(d_{test}, d_{test})}
            \end{split}
        \label{eq:redundancy}
        \end{equation}

    \subsection{Benchmark Settings}
    
        \subsubsection{Hardware}
        \label{sec:hardware}
    
            All experiments were conducted on a computer equipped with an Intel Core i9-9900KF CPU (3.60 GHz), a NVIDIA RTX 4090 GPU (24GB), and 32GB of RAM, provided by the University of the Balearic Islands. The hardware specifications were considered sufficient to perform all necessary computations within a reasonable timeframe.

        \subsubsection{Models}
        \label{sec:models}
    
            Two deep learning models were selected for the experiments, to include different architectural choices: Swin Transformer~\cite{liu2021swin}, and ConvNext~\cite{liu2022convnet}. The two of them are fairly recent and have demonstrated excelling results in the image classification task, with results above 80\% accuracy on the ImageNet dataset.
            

            The Swin Transformer is a hierarchical vision transformer that processes images using shifted windows, improving efficiency and scalability. It constructs multi-scale feature maps by splitting images into patches and merging them at deeper layers, enabling strong performance in object detection and segmentation. The shifted window approach balances local self-attention with cross-window connections, allowing Swin Transformer to excel in various vision tasks.  
            

            ConvNeXt revisits ConvNets by integrating transformer-inspired design choices, such as large kernel sizes and inverted bottlenecks, into a ResNet-based architecture. This modernization enables ConvNeXt to surpass transformer models like Swin Transformer in tasks such as object detection and segmentation while maintaining the efficiency of ConvNets. It demonstrates that well-designed ConvNets remain highly competitive for modern vision tasks.

        \subsubsection{Training Process}
    
            We conducted five training sessions for each dataset, following a cross-validation procedure. This resulted in a total of 240 training sessions ($24 \text{ datasets} \times 2 \text{ networks} \times 5 \text{ cross validations} = 240$). The models were implemented using the PyTorch framework, leveraging pre-trained weights from ImageNet. Data augmentation techniques were applied uniformly, including random horizontal flipping, rotation, translation, scaling, and adjustments to brightness and contrast. Each model was trained for a maximum of 20 epochs, with early stopping applied when validation accuracy did not improve by at least 1\% for 5 consecutive epochs. Furthermore, since some of the datasets used for training were unbalanced, we weighted the loss function (cross entropy loss) to account for class imbalance.
    
         \subsubsection{Testing Process}
    
            Every training was evaluated on every other dataset. In total, $240 \text{ trainings} \times 24 \text{ datasets} = 5,760$ evaluations were conducted. For fairness, since not all seven expression classes were present in all datasets, only those present in the training were used for its testing on the remaining datasets. 
    
            After computing all evaluations, we measured three key characteristics for each dataset: \textit{Local Similarity}, \textit{Global Similarity}, and \textit{Paired Similarity}, described in Section \ref{sec:metrics}. As measure of performance $\hat\theta$, we used the F1 score metric, a very common metric for image classification tasks that stays fair for unbalanced problems, conversely to accuracy~\cite{branco2016survey}. 
            




\section{Results}

    In this section, we present the results obtained to answer the primary and secondary research questions outlined in Section \ref{sec:rqs}. Each subsection corresponds to a specific research question.

    \subsection{RQ1: Datasets Exploration}

\begin{figure}[h]
             \centering
             \includegraphics[width=\columnwidth]{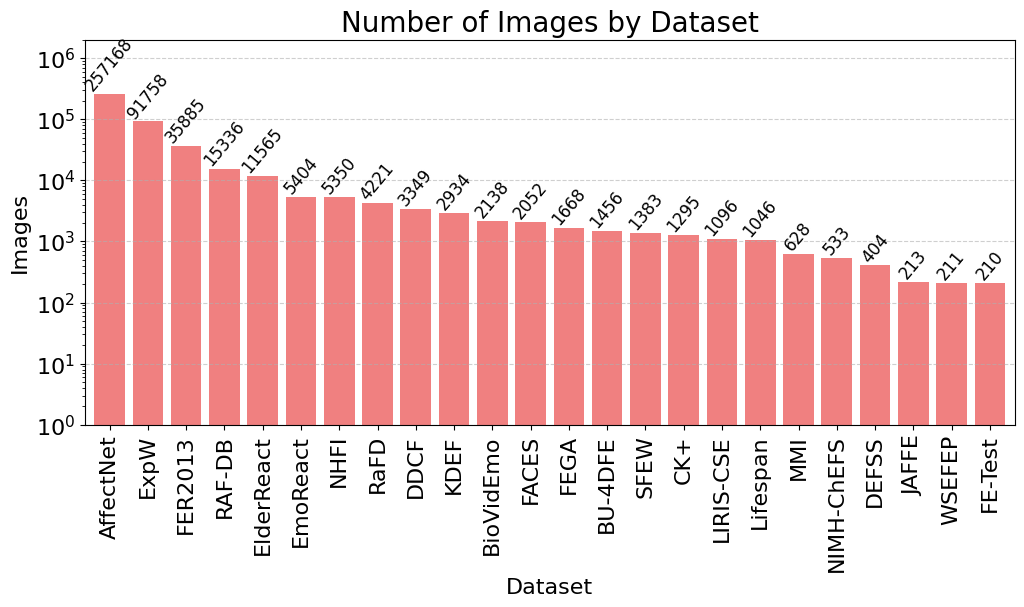}
             \caption{Number of images per dataset, in logarithmic scale.}
            \label{fig:dataset-count}
        \end{figure}

        To address the first primary research question, we examine the key characteristics of the collected datasets, considering various aspects. Figure~\ref{fig:dataset-count} shows the number of images in each dataset after applying the normalization process described in Section \ref{sec:collection-normalization}.

        \subsubsection{RQ1.1: Collection Process}
            
            Among the collected datasets, nine--AffectNet, ExpW, FER2013, RAF-DB, NHFI, FE-Test, SFEW, ElderReact, and EmoReact--were produced automatically from web data extraction. Of these, AffectNet, ExpW, FER2013, RAF-DB, and NHFI were collected using various query strings and later annotated by human labelers. Because they share common image sources, duplicate images may appear across these datasets. Additionally, due to their automated nature, these datasets tend to be large but may contain labeling errors, duplicate images, and non-facial images~\cite{mejia-escobar2023towards, liu2022uncertain, ijimai}. In contrast, FE-Test, ElderReact, and EmoReact were manually curated from online sources. FE-Test was hand-picked to create a small test set for evaluating models trained on other datasets. ElderReact and EmoReact were sourced from the ``React" YouTube channel\footnote{\url{https://www.youtube.com/user/React}}, then clipped and annotated. Notably, SFEW is a static version of the AFEW dataset, created by extracting frames from movie clips using subtitles to identify specific expressions.
            
            The remaining datasets were captured in controlled environments using one or more cameras. This setup allows for greater control over factors such as face orientation, gaze, lighting conditions, and other key variables. Additionally, since these datasets feature posed expressions, the labels correspond to the intended expressions of the actors, reducing the likelihood of labeling errors compared to automatically collected datasets. These datasets also tend to include demographic annotations (e.g., age, gender, and ethnicity) and offer multiple expression samples per individual, making them suitable for cross-subject experiments.

\begin{figure}[h]
                \captionsetup[subfigure]{justification=centering}
                 \centering
                 \begin{subfigure}[b]{.49\columnwidth}
                     \centering
                     \includegraphics[width=\columnwidth]{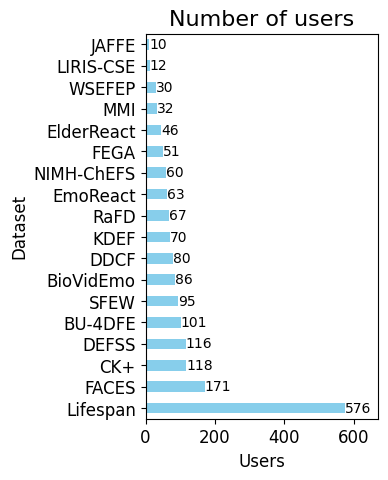}
                     \caption{ }
                    \label{fig:dataset-users}
                 \end{subfigure}
                 \begin{subfigure}[b]{.49\columnwidth}
                     \centering
                     \includegraphics[width=\columnwidth]{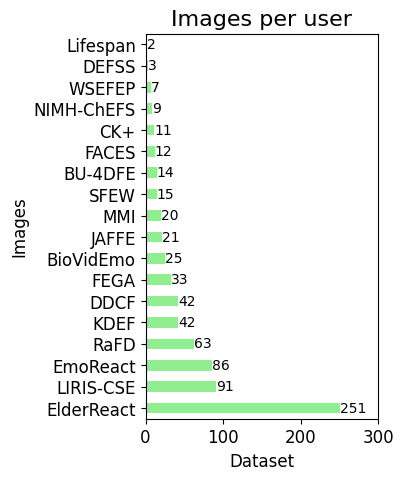}
                     \caption{ }
                    \label{fig:dataset-images-users}
                 \end{subfigure}
                \caption{(a) Number of users by dataset. (b) Average number of images per user by dataset.}
                \label{fig:times}
            \end{figure}

            Figure~\ref{fig:dataset-users} presents the number of individuals included in each dataset, while Figure~\ref{fig:dataset-images-users} shows the average number of images per user. Note that automatically collected datasets lack user demographic labels, and hence were excluded from the figures. Both the number of users and the number of images per user vary significantly across datasets, although datasets with a large number of users generally contain fewer images per user, e.g. Lifespan and DDCF.

            Among the collected datasets, only four--ElderReact, EmoReact, LIRIS-CSE, and BioVidEmo--were explicitly designed to capture spontaneous expressions. These datasets were recorded in controlled environments while attempting to elicit natural reactions using multimedia stimuli. In contrast, Internet-sourced datasets such as AffectNet and FER2013 may contain a mix of natural and posed expressions, whereas the remaining datasets feature posed expressions.
    
        \subsubsection{RQ1.2: Demographic Characteristics}

\begin{figure}[h]
                 \centering
                 \includegraphics[width=\columnwidth]{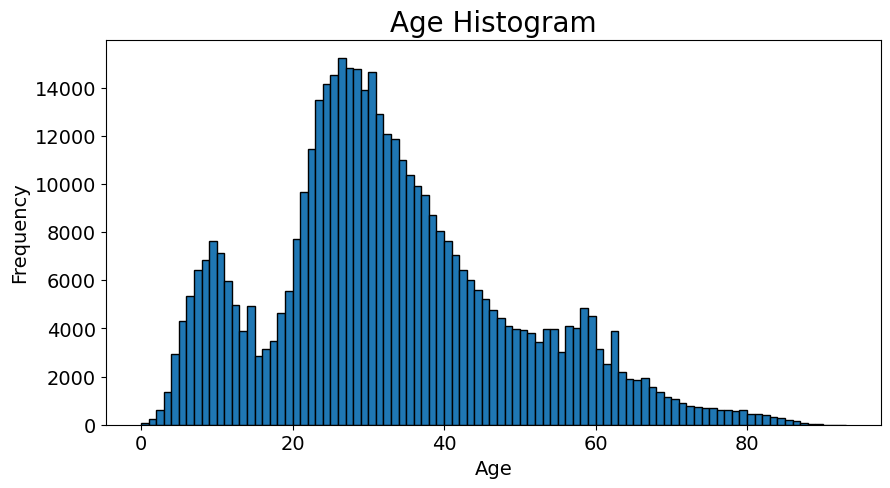}
                 \caption{Age histogram of the combined datasets.}
                \label{fig:dataset-age}
            \end{figure}

            Figure~\ref{fig:dataset-age} presents the age histogram across all datasets. A total of 55\% of the images feature individuals between 20 and 40 years old, while only 21\%, 17\% and 7\% correspond to individuals between 40 and 60, under 18, and over 60 years old, respectively. This indicates a significant age-related bias, with a much higher representation of adults compared to children and older adults. The availability of images for elderly individuals is particularly limited and decreases further with increasing age. Interestingly, the data appears to form three distinct clusters corresponding to different age groups: around 10 years old (children), 30 years old (adults), and 60 years old (elderly).

\begin{figure}[h]
                 \centering
                 \includegraphics[width=\columnwidth]{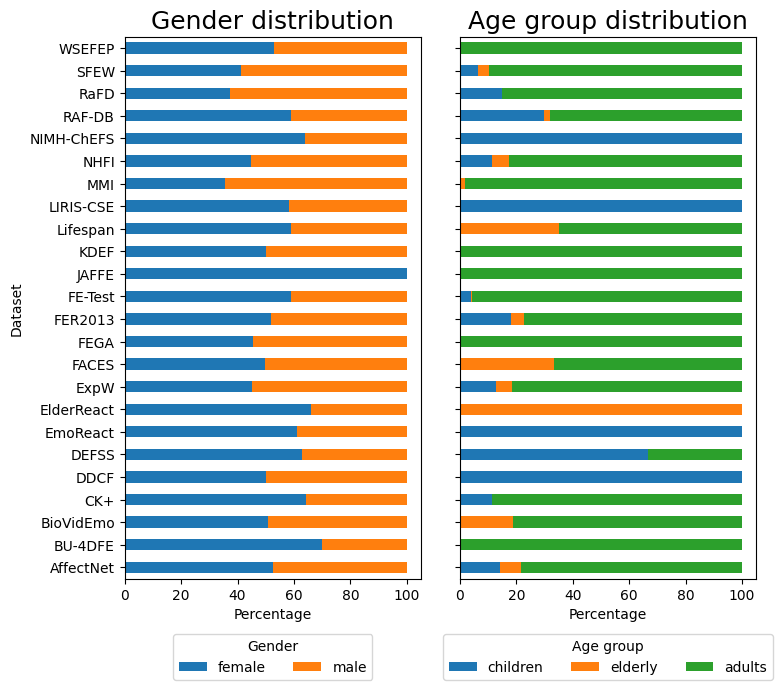}
                 \caption{(Left) Gender distribution across datasets. (Right) Age group distribution across datasets.}
                \label{fig:dataset-gender-age-group}
            \end{figure}

            A more detailed analysis of age distribution per dataset reveals significant variability, as illustrated in the right plot of Figure~\ref{fig:dataset-gender-age-group}. Large-scale automated datasets--AffectNet, FER2013, RAF-DB, SFEW, ExpW, and NHFI--contain images spanning all age groups but tend to be biased toward the adult category. Conversely, manually curated datasets often focus on specific age groups. For instance, LIRIS-CSE, EmoReact, DDCF, DEFSS, and NIMH-ChEFS primarily include children, while ElderReact is dedicated to elderly individuals. Other datasets, such as WSEFEP, MMI, KDEF, JAFFE, FEGA, and BU-4DFE, predominantly feature adult subjects. Notably, two manually curated datasets, Lifespan and FACES, cover a broader age range, from young adults to older individuals. These datasets were specifically designed to explore the impact of aging on facial expression recognition, though they lack images of children, preventing full-spectrum age coverage. Additionally, some acted datasets primarily consist of adult subjects but include a small proportion of children (RaFD, DEFSS, and CK+) and elderly individuals (BioVidEmo).

            The left plot of Figure~\ref{fig:dataset-gender-age-group} illustrates the gender distribution across datasets. In most cases, the difference between male and female representation is less than 20\%. However, some datasets show a stronger gender imbalance. Specifically, RAFD and MMI have a male predominance, whereas JAFFE, BU-4DFE, ElderReact, NIMH-ChEFS, DEFSS, CK+, and EmoReact have a higher proportion of female images. This suggests a slight overall bias toward female representation, with JAFFE and BU-4DFE exhibiting the most pronounced imbalance (100\% and 70\% female subjects, respectively).


            Regarding ethnicity, only WSEFEP, JAFFE, DEFSS, Lifespan, FEGA, FACES, KDEF, DDCF, and RaFD include explicit ethnicity labels. There is a strong predominance of the Caucasian ethnicity (also referred to as ``White" in some datasets), although some datasets provide more diversity. For example, RaFD includes images of Moroccan individuals, JAFFE features Japanese subjects, and Lifespan includes images from individuals categorized as ``Black" and Indian. For datasets without ethnicity labels, some insights can be gathered from their respective publications. For example, in NIMH-ChEFS, most participants are Caucasian, with four girls and one boy identified as non-Caucasian. The BU-4DFE dataset contains diverse ethnic representation: Asian (28 participants), Black (8), Hispanic/Latino (3), and White (62). In CK+, 81\% of subjects are Euro-American, 13\% Afro-American, and 6\% from other groups. The MMI dataset comprises 10 European, 3 South American, and 12 Asian participants. The remaining datasets neither include ethnicity labels nor provide relevant information in their documentation.

    
        \subsubsection{RQ1.3: Classes}

            The most commonly found classes across the collected datasets correspond to the six basic expressions~\cite{ekmanuniversal} plus the neutral expression. However, some datasets also include additional expressions. For instance, the ``contempt" expression appears in AffectNet, NHFI, RaFD, and CK+; ``pleased" is present in DDCF; ``amusement" in BioVidEmo; and ``joy" in WSEFEP. Moreover, EmoReact provides labels for ``curiosity", ``uncertainty", ``excitement", and ``frustration"; Lifespan includes ``annoyance", ``grumpiness", and ``profile"; while MMI contains ``scream", ``boredom", and ``sleepiness". Notably, some of these expressions, such as ``amusement" and ``joy", are closely related to a basic expression--in this case, ``happiness".
            
\begin{figure}[h]
                 \centering
                 \includegraphics[width=\columnwidth]{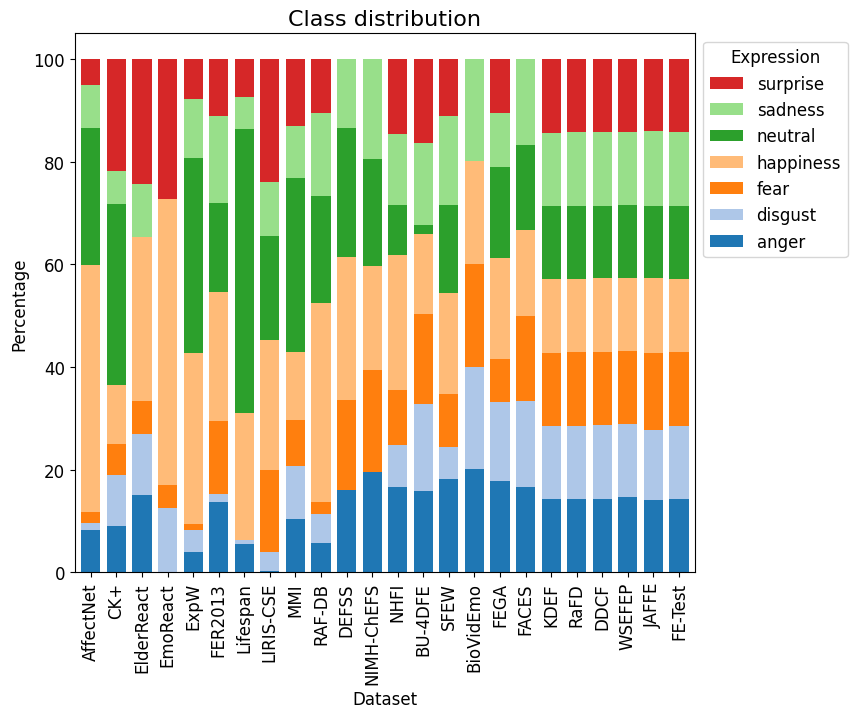}
                 \caption{Class distribution across datasets.}
                \label{fig:dataset-classes-percentage}
            \end{figure}

            Although most datasets include labels for the six basic expressions plus the neutral one, their distributions are often highly imbalanced. Figure~\ref{fig:dataset-classes-percentage} presents the class distribution across datasets. As shown, ``happiness" and ``neutral" are frequently the two most prevalent classes, whereas ``fear" and ``disgust" tend to be underrepresented in many datasets, including AffectNet, EmoReact, ExpW, Lifespan, and RAF-DB. In smaller datasets, minority classes may contain an insufficient number of images for effective training. For example, the Lifespan dataset includes only 7 images for the ``disgust" class.
            
            Additionally, some datasets lack certain expressions: LIRIS-CSE, FACES, BU-4DFE, ElderReact, and MMI do not include the ``neutral" expression; Lifespan lacks ``fear"; DEFSS and NIMH-ChEFS omit ``disgust" and ``surprise"; BioVidEmo excludes ``surprise" and ``neutral"; and EmoReact does not contain ``anger", ``neutral", or ``sadness".  

        \subsubsection{RQ1.4: Data Characteristics}

            Table~\ref{tab:data} summarizes the characteristics of the collected datasets, including format (image or video), resolution, color, whether the images are cropped to the face, and whether they include variations in perspective, background, and illumination conditions.

            \begin{table}[h]
            \caption{Data characteristics of the collected datasets}
            \label{tab:data}
            \centering
            \setlength{\tabcolsep}{3pt}
            \resizebox{\columnwidth}{!}{
            \begin{tabular}{l|llll|lll}
                \toprule
                \multicolumn{1}{c|}{\multirow{2}{*}{\textbf{Dataset}}} & \multicolumn{4}{c|}{\textbf{Data}} & \multicolumn{3}{c}{\textbf{Variations}} \\
                \multicolumn{1}{c|}{} & \textbf{Format} & \textbf{Resolution} & \textbf{Color} & \textbf{F. crop} & \textbf{Persp.} & \textbf{Backgr.} & \textbf{Illum.} \\
                \midrule
                \textbf{AffectNet} & Image & Variable & RGB & Yes & Yes & Yes & Yes \\
                \textbf{BioVidEmo} & Video & $1388 \times   1038$ & RGB & No & Yes & No & No \\
                \textbf{BU-4DFE} & Video & $312 \times 418$ & RGB & Yes & No & No & No \\
                \textbf{CK+} & Video & $640 \times   490$ & Grayscale & Yes & No & No & Yes \\
                \textbf{DDCF} & Image & $900 \times 900$ & RGB & Yes & Yes & No & Yes \\
                \textbf{DEFSS} & Image & $850 \times   947$ & RGB & Yes & No & No & No \\
                \textbf{ElderReact} & Video & $1280 \times 720$ & RGB & No & Yes & Yes & No \\
                \textbf{EmoReact} & Video & $1280 \times   720$ & RGB & No & Yes & Yes & No \\
                \textbf{ExpW} & Image & Variable & RGB & Both & Yes & Yes & Yes \\
                \textbf{FACES} & Image & $2835 \times   3543$ & RGB & Yes & No & No & No \\
                \textbf{FEGA} & Image & $640 \times 480$ & RGB & No & No & Yes & No \\
                \textbf{FER2013} & Image & $48 \times 48$ & Grayscale & Yes & Yes & Yes & Yes \\
                \textbf{FE-Test} & Image & $150 \times 150$ & Grayscale & Yes & Yes & Yes & Yes \\
                \textbf{JAFFE} & Image & $256 \times   256$ & Grayscale & Yes & No & No & No \\
                \textbf{KDEF} & Image & $562 \times 762$ & RGB & Yes & Yes & No & No \\
                \textbf{Lifespan} & Image & Variable & RGB & Yes & No & No & No \\
                \textbf{LIRIS-CSE} & Video & $800 \times 600$ & RGB & No & Yes & Yes & No \\
                \textbf{MMI} & Video & $720 \times   576$ & RGB & Yes & Yes & No & No \\
                \textbf{NHFI} & Image & $224 \times 224$ & Grayscale & Yes & Yes & Yes & Yes \\
                \textbf{NIMH-ChEFS} & Image & $1960 \times   3008$ & RGB & Yes & No & No & No \\
                \textbf{RaFD} & Image & $681 \times 1024$ & RGB & Yes & Yes & No & No \\
                \textbf{RAF-DB} & Image & $100 \times   100$ & RGB & Yes & Yes & Yes & Yes \\
                \textbf{SFEW} & Image & $720 \times 576$ & RGB & No & Yes & Yes & Yes \\
                \textbf{WSEFEP} & Image & $1725 \times   1168$ & RGB & Yes & No & No & No \\
                \bottomrule
            \end{tabular}
            }
            \end{table}

            Only seven datasets are available in video format, all of which were collected under controlled conditions (none were automatically sourced from the Internet). As a result, these video datasets contain a limited number of clips, posing challenges for expression recognition based on dynamic sequences rather than static images. Notably, SFEW is the static version of AFEW, which is originally in video format. The datasets exhibit a wide range of resolutions. With the exception of AffectNet, ExpW, and Lifespan, they contain images in a fixed resolution, ranging from low resolutions, such as FER2013 ($48 \times 48$) and RAF-DB ($100 \times 100$), to high resolutions, such as FACES ($2835 \times 3543$) and NIMH-ChEFS ($1960 \times 3008$). Most datasets provide images in RGB, except for CK+, FER2013, FE-Test, JAFFE, and NHFI, which are in grayscale. Additionally, most datasets contain face-centered images, except for BioVidEmo, ElderReact, EmoReact, FEGA, LIRIS-CSE, and SFEW, which provide uncropped images. Notably, ExpW includes uncropped images along with bounding box annotations for cropping.

            Several datasets include variations in facial perspective, background, and illumination conditions. As expected, datasets collected from the Internet--AffectNet, ExpW, RAF-DB, SFEW, NHFI, FE-Test, and FER2013--naturally exhibit these variations, as the images were captured in diverse contexts. Video-format datasets generally offer variations in facial perspective due to head movements throughout the videos, as seen in BioVidEmo, ElderReact, EmoReact, and LIRIS-CSE. Additionally, MMI, DDCF, KDEF, and RaFD deliberately incorporated different facial viewpoints during data collection. The majority of datasets created using actors maintained consistent background and illumination conditions across all images. This was the case for BioVidEmo, BU-4DFE, DEFSS, FACES, JAFFE, KDEF, Lifespan, MMI, NIMH-ChEFS, RaFD, and WSEFEP. In contrast, DDCF and CK+ included variations in illumination intensity, while ElderReact, EmoReact, FEGA, and LIRIS-CSE featured recordings in different scenarios. Notably, DDCF’s actors posed wearing a black beanie, while BioVidEmo’s actors wore an EEG cap.

    \subsection{RQ2: Benchmark}

        In this section, we present the results of the benchmarking process conducted on the collected datasets to address RQ2 and its secondary research questions.

        \subsubsection{RQ2.1: Dataset Difficulty}

\begin{figure}[h]
                 \centering
                 \includegraphics[width=\columnwidth]{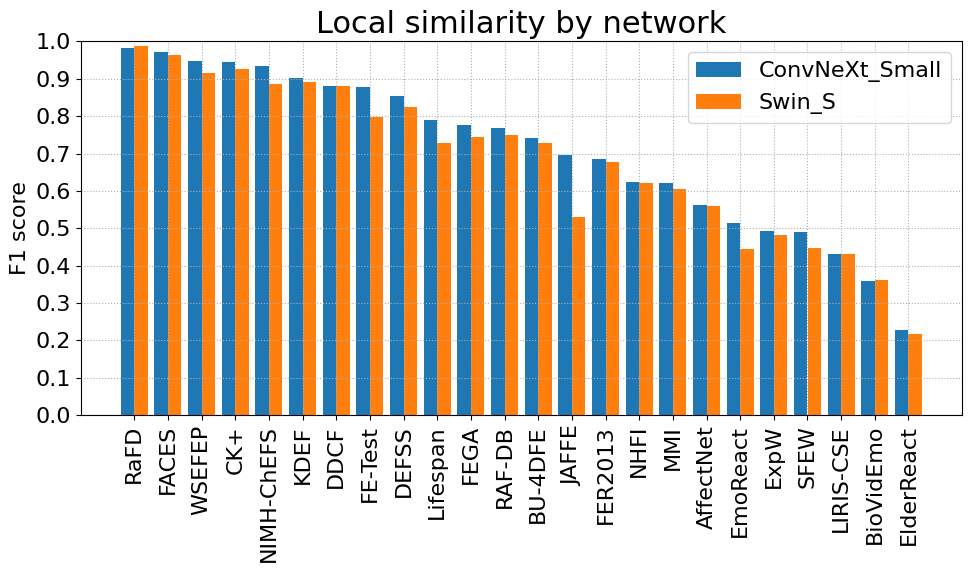}
                 \caption{Local similarity computed per network.}
                \label{fig:local-similarity}
            \end{figure}

            The numerical results for the Local Similarity metric are shown in the second column of Table~\ref{tab:local-global}. Figure~\ref{fig:local-similarity} presents the results for this metric, broken down by network. This metric evaluates the performance of networks trained and tested on the same dataset, providing insights into the relative difficulty of each dataset.

            \begin{table}[h]
            \caption{Local and Global Similarity of the datasets}
            \label{tab:local-global}
            \centering
            \begin{tabular}{l|ll}
                \toprule
                \textbf{Dataset} & \textbf{Local Similarity} & \textbf{Global Similarity}
                \\
                \midrule
                AffectNet & 0.5622 & 0.6095 \\
                BioVidEmo & 0.3603 & 0.2030 \\
                BU-4DFE & 0.7348 & 0.3981 \\
                CK+ & 0.9355 & 0.4121 \\
                DDCF & 0.8797 & 0.3978 \\
                DEFSS & 0.8395 & 0.3058 \\
                ElderReact & 0.2224 & 0.1667 \\
                EmoReact & 0.4806 & 0.0994 \\
                ExpW & 0.4882 & 0.4782 \\
                FACES & 0.9681 & 0.3707 \\
                FEGA & 0.7606 & 0.4017 \\
                FER2013 & 0.6807 & 0.5008 \\
                FE-Test & 0.8384 & 0.3368 \\
                JAFFE & 0.6129 & 0.2686 \\
                KDEF & 0.8966 & 0.4551 \\
                Lifespan & 0.7587 & 0.1769 \\
                LIRIS-CSE & 0.4318 & 0.1934 \\
                MMI & 0.6130 & 0.3977 \\
                NHFI & 0.6229 & 0.4773 \\
                NIMH-ChEFS & 0.9107 & 0.2916 \\
                RAF-DB & 0.7578 & 0.4601 \\
                RaFD & 0.9849 & 0.4539 \\
                SFEW & 0.4681 & 0.2795 \\
                WSEFEP & 0.9314 & 0.3695 \\
                \bottomrule
            \end{tabular}
            \end{table}

            The highest Local Similarity values were observed in RaFD, FACES, WSEFEP, and NIMH-ChEFS, all exceeding 0.9. These datasets were collected under controlled conditions, likely resulting in fewer variations in images and, consequently, making the recognition task easier. Among the datasets collected automatically from the Internet, RAF-DB and FER2013 achieved scores above 0.7, while NHFI, AffectNet, and ExpW had lower scores. The lowest scores were recorded for ElderReact (0.2224), BioVidEmo (0.3603), LIRIS-CSE (0.4318), and SFEW (0.4681). Interestingly, all of these datasets were originally in video format and were converted into image sets via frame sampling. However, two other video-format datasets, CK+ and MMI, performed significantly better, with values above 0.6.

            The highest variability in results across networks was observed in JAFFE and FE-Test, both containing fewer than 400 samples. In contrast, other similarly sized datasets, such as WSEFEP and DEFSS, did not exhibit such variations. Among cross-validation iterations, the highest difference was on JAFFE by far, followed by Lifespan, and FEGA.

        \subsubsection{RQ2.2: Dataset Generalization}

\begin{figure}[h]
                 \centering
                 \includegraphics[width=\columnwidth]{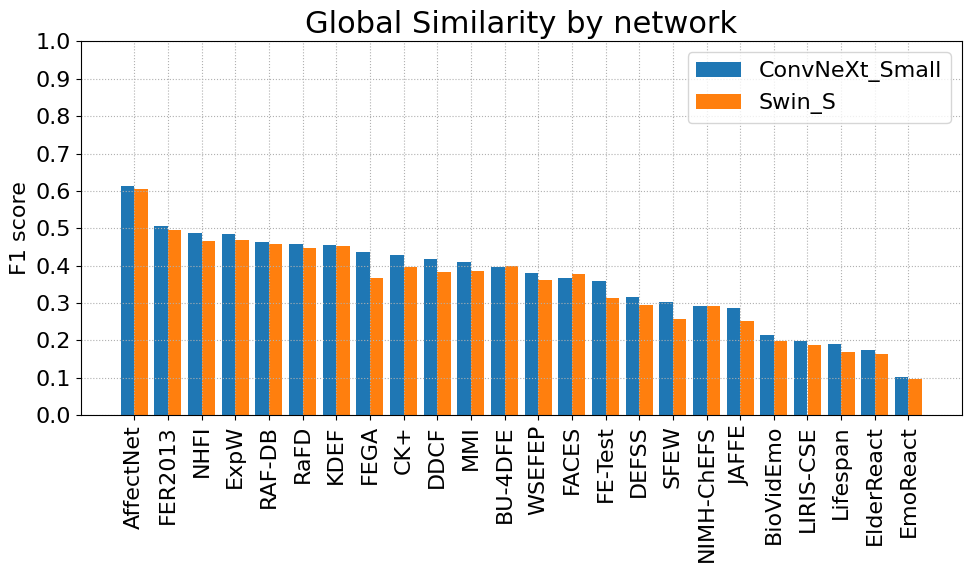}
                 \caption{Global similarity computed per network.}
                \label{fig:global-similarity}
            \end{figure}

            The results for the Global Similarity metric are shown in the third column of Table~\ref{tab:local-global}. Figure~\ref{fig:global-similarity} illustrates the results, broken down by network. Unlike Local Similarity, which evaluates networks on the same dataset they were trained on, this metric measures performance when networks are tested on all datasets except the one used for training, thereby assessing generalization capabilities.

            The highest Global Similarity score was achieved by AffectNet (0.6095), which was 0.1 higher than the second-best dataset, FER2013 (0.5008). Other high-performing datasets included NHFI (0.4773), ExpW (0.4782), and RAF-DB (0.4601), all of which were collected from the Internet and are among the largest datasets. In contrast, datasets collected under controlled conditions performed worse in terms of generalization. The lowest scores were obtained by EmoReact (0.0994), followed by ElderReact (0.1667), Lifespan (0.1769), and LIRIS-CSE (0.1934). Once again, video-format datasets occupied the lowest rankings, with the exception of CK+ and MMI. The greatest variation in network performance was observed in FEGA and FE-Test.

        \subsubsection{RQ2.3: Dataset Redundancy}

\begin{figure*}[h]
                 \centering
                 \includegraphics[width=.75\textwidth]{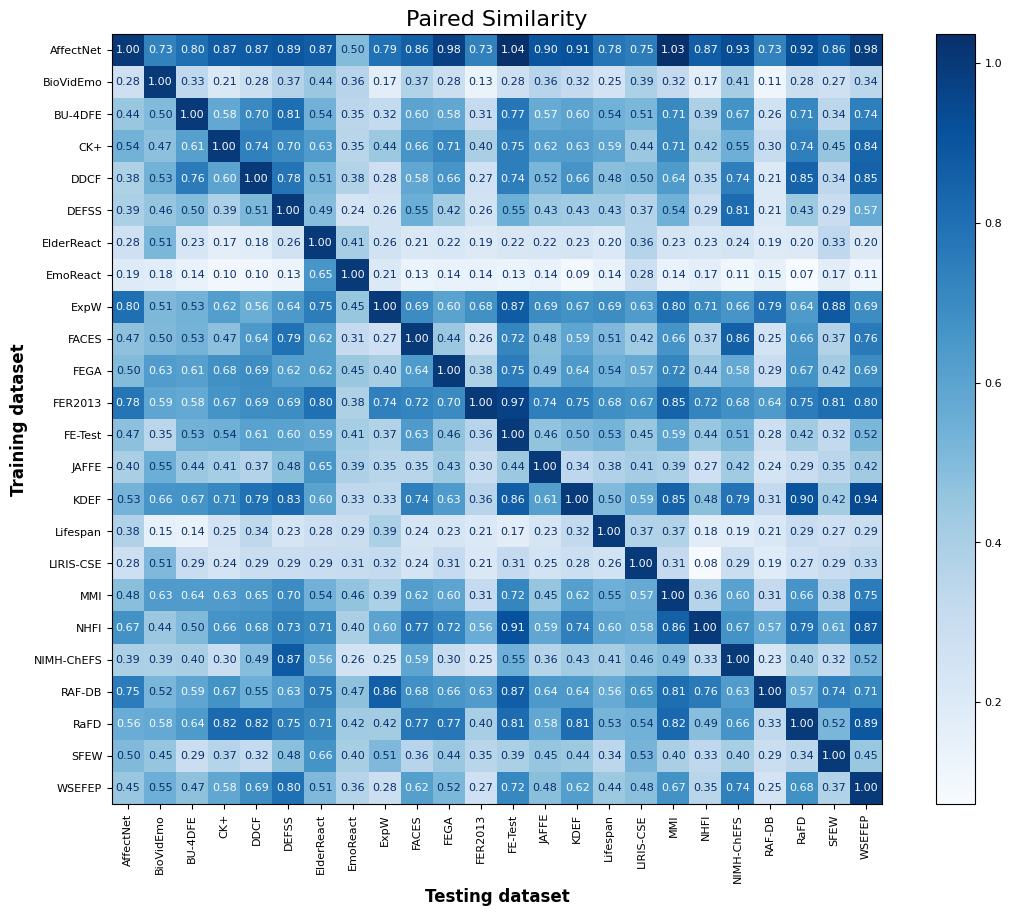}
                 \caption{Paired similarity of the collected datasets.}
                \label{fig:paired-similarity}
            \end{figure*}

            Figure~\ref{fig:paired-similarity} presents the Paired Similarity of the datasets, where each training dataset is evaluated against all other datasets and normalized by its Local Similarity. The results are displayed in matrix form, where rows represent the training dataset and columns represent the testing dataset. The diagonal entries are always equal to one, as they correspond to the paired similarity of each dataset with itself. However, values greater than one indicate cases where training on one dataset results in better performance on another dataset than on itself. This metric provides insights into how much information from one dataset can be learned from another.

            There were two cases where the Paired Similarity exceeded 1: training on AffectNet resulted in better performance on FE-Test, and MMI. This suggests that AffectNet may provide more useful training features than those two datasets themselves, likely due to factors such as dataset size and similarity. Additionally, AffectNet achieved Paired Similarity scores above 0.90 for FEGA, WSEFEP, JAFFE, KDEF, NIMH-ChEFS, and RaFD; FER2013 for FE-Test; KDEF for RaFD and WSEFEP; and NHFI for FE-Test. These findings align with the Global Similarity results discussed in the previous section, where these datasets--particularly AffectNet--demonstrated strong generalization capabilities.

            Conversely, EmoReact, ElderReact, BioVidEmo, Lifespan, and LIRIS-CSE consistently exhibited the lowest scores, both as training and testing datasets, which aligns with their poor Global Similarity performance. Notably, FE-Test generally achieved high Paired Similarity when used as testing dataset but had lower values when used for training.

\section{Discussion}

    The exploration and benchmarking of the collected datasets revealed common characteristics, strengths, and weaknesses. In this section, we use the obtained results to explore these aspects in depth and list recommendations for future studies in the facial expression recognition field.

    \subsection{Collection Process, Demographic Biases and Expression Distribution}

        The main difference between datasets stems from their collection process: automatic collection from the Internet or manual recording with actors in a controlled setup. This difference significantly impacts the characteristics of the images in the dataset. Automatic datasets tend to be larger and exhibit greater variation in illumination conditions, backgrounds, face perspectives, and demographic diversity. However, unlike manually recorded datasets, they lack demographic information such as age, gender, and ethnicity and may contain lower-quality expression annotations since they were labeled a posteriori, meaning that the labels may be influenced by labelers' subjectivity. Another key distinction between datasets is the nature of the expressions they contain, either acted or spontaneous. Only four datasets--ElderReact, EmoReact, LIRIS-CSE, and BioVidEmo--offer spontaneous expressions, elicited by showing videos to participants. Nevertheless, these datasets performed poorly across all three metrics, indicating the increased difficulty of recognizing spontaneous expressions and their unsuitability for expression recognition from static images.
    
        Datasets lacking demographic information were automatically labeled using computational tools. This analysis revealed a significant age imbalance in all datasets, with a predominance of adult samples over children and elderly individuals. Since crucial facial differences exist between these age groups that affect facial expression recognition~\cite{fölster2014facial}, the underrepresentation of minority groups may lead to biased training. The same issue arises with ethnicity, where datasets are overwhelmingly dominated by Caucasian (or white) participants. On the other hand, gender distribution was generally balanced across datasets.
        
        While custom datasets typically maintain a balanced number of samples per expression, automatically collected datasets tend to be heavily biased toward happiness and neutral expressions, as these are more commonly found on the Internet and typically easier to distinguish from other expressions. Addressing this imbalance is crucial to avoid biases in training, with one straightforward approach being the use of a weighted loss function, as implemented in this study.

    \subsection{Benchmarking with Similarity Metrics}

        Numerous metrics exist in information theory to assess the similarity of datasets, such as entropy-related metrics. However, in this study, our goal was not to compare datasets based on image similarity but rather in terms of the knowledge deep learning models can extract from them. In this regard, the three proposed metrics--Local, Global, and Paired Similarity--offer valuable insights when given a performance metric, a set of datasets, and a set of networks. Local Similarity helped identify easy and difficult datasets, often influenced by dataset size, variability, and annotation quality. Global Similarity pinpointed datasets that enabled strong generalization across unseen data or, conversely, exhibited high specificity. Paired Similarity highlighted specific dataset pairs that were either very similar or highly dissimilar in terms of the knowledge networks could transfer between them.

        Despite being automatically collected and containing potential label errors, duplicated images, and missing demographic information~\cite{ijimai, mejia-escobar2023towards, liu2022uncertain}, the largest datasets demonstrated superior generalization. These datasets consistently ranked among the top performers in the Global Similarity metric, likely due to their high variability in image sources and conditions. However, networks struggled to achieve high performance on these datasets, as reflected in their low Local Similarity scores. This suggests that the very factor contributing to their strong generalization--their diversity--also makes them particularly challenging test datasets. Despite these difficulties, some automatic datasets even outperformed models trained and tested on the same dataset. Notably, AffectNet, the largest dataset in the benchmark, achieved the best results on FE-Test, and MMI, even surpassing models trained directly on those datasets. These findings suggest that large-scale, diverse datasets are best suited for real-life, uncontrolled scenarios. Conversely, datasets recorded with actors in controlled settings exhibited higher Local Similarity scores, likely due to their lower variability, but at the cost of reduced Global Similarity and weaker generalization.
        
        The size of a dataset significantly influences training outcomes, even when using the same network. Greater variations were observed in cross-validation iterations when training on smaller datasets such as JAFFE, and FE-Test, potentially leading to overfitting. Nevertheless, FE-Test, and WSEFEP performed reasonably well on average. Additionally, images extracted from videos yielded worse results across all metrics compared to static photos. This may be due to limited variation between consecutive frames and the inclusion of low-intensity expressions or transitional frames, which are harder to classify than peak-intensity expressions in static image datasets.

        Analyzing the three similarity metrics--Local, Global, and Paired Similarity--allowed us to identify the overall weakest-performing datasets. The most evident cases are ElderReact, EmoReact, LIRIS-CSE, and BioVidEmo, which performed poorly across all metrics. ElderReact had the lowest Local Similarity score ($\approx 0.2$) and a similarly low Global Similarity score, indicating that networks failed to learn meaningful distinctions between its facial expressions and that training on this dataset did not generalize well to others. EmoReact had the lowest Global Similarity ($\approx 0.1$), while LIRIS-CSE and BioVidEmo also exhibited poor performance ($\text{Local Similarity} < 0.5$, $\text{Global Similarity} < 0.2$). Notably, these results were close to those of a random classifier, which would achieve a performance of approximately 0.14 in a 7-class classification task.  The poor results of these datasets may be attributed to their original video format, where expressions do not persist throughout the entire clip. Consequently, after frame sampling, some images may be labeled with an expression they do not truly depict, increasing the difficulty of expression recognition from still images.

    \subsection{Recommendations for Future FER Studies}

        Based on our dataset exploration and benchmarking results, we provide the following recommendations for future FER studies:
    
        \begin{itemize}
        
            \item \textbf{Avoid small datasets.} Training on datasets with too few samples may lead to overfitting and poor generalization. Instead, use datasets with at least 500 samples.
            
            \item \textbf{Prioritize size over quality for generalization.} The best generalization results (Global Similarity) were achieved with large datasets like AffectNet and FER2013, despite their known labeling issues~\cite{mejia-escobar2023towards, liu2022uncertain}.
            
            \item \textbf{Avoid using sampled videos.} Datasets derived from videos generally contain variable-intensity expressions, making them unsuitable for single-image expression recognition, as evidenced by the poor performance of ElderReact, EmoReact, BioVidEmo, and LIRIS-CSE.
            
            \item \textbf{Address class imbalances.} Severe expression count imbalances in most datasets necessitate rebalancing techniques or weighted loss functions to mitigate bias.
            
            \item \textbf{Ensure demographic diversity.} Most datasets are heavily imbalanced in terms of age and ethnicity. The target use case should be examined beforehand to avoid deployment biases. Target users of the system should have a fair representation in the dataset used.
            
            \item \textbf{AffectNet is a strong choice for many applications.} Its size allows for superior generalization, achieving strong results even on datasets such as FACES, RaFD and WSEFEP, recorded under controlled conditions.
            
            \item \textbf{Consider dataset similarity when combining multiple datasets.} Paired Similarity analysis revealed that some datasets are redundant. For instance, training on AffectNet alone yields better results on MMI than training on MMI itself. Therefore, study Paired Similarity before including multiple datasets in a training set.
            
        \end{itemize}

    \subsection{Recommendations for New FER Datasets}

        For studies aiming to construct new FER datasets, we recommend:
    
        \begin{itemize}
        
            \item \textbf{Opt for large-scale data collection.} Controlled recordings with actors can be useful for specific studies but often limit dataset size. Automatic collection methods may be simpler and more effective for acquiring large datasets.
            
            \item \textbf{Incorporate variations in lighting, settings, and perspectives.} Limited variation increases Local Similarity but hinders generalization.
            
            \item \textbf{Ensure demographic diversity if using actors.} While gender is often balanced, age and ethnicity are frequently underrepresented.
            
            \item \textbf{Include rich metadata.} Labels for age, gender, ethnicity, and other characteristics (e.g., glasses, facial hair) enable more nuanced studies.
            
            \item \textbf{Eliminate duplicates in automatically collected datasets.} Duplicate images can lead to overfitting and unfair performance evaluations if they appear in both training and test sets.
            
        \end{itemize}

\section{Conclusion}

    This study presents a comprehensive evaluation of 24 widely used FER datasets to uncover their intrinsic characteristics and determine their suitability for training deep learning models. By employing a rigorous normalization process and enriching the datasets with additional automatic demographic annotations, we have been able to examine key factors such as demographic biases, class imbalances, and data variability. Additionally, three metrics were introduced--Local, Global, and Paired Similarity--that allowed for a detailed quantitative analysis of each dataset’s learning difficulty, generalization capability, and the extent of knowledge transfer between datasets, offering a fair and thorough benchmark of current FER datasets.
    
    Despite containing label errors, duplicates, and missing demographic details, large-scale automatically collected datasets demonstrate strong generalization (high Global Similarity) due to their high variability. However, this same diversity poses challenges for deep learning models, leading to lower Local Similarity scores. Conversely, datasets recorded with actors in controlled settings yield higher Local Similarity due to consistent conditions, yet they exhibit reduced Global Similarity and weaker generalization, making them less effective in real-life, uncontrolled scenarios. Smaller datasets, such as JAFFE, and FE-Test, can lead to overfitting, while images extracted from videos generally perform worse than static photos because they often include transitional frames or low-intensity expressions. Notably, datasets like ElderReact, EmoReact, LIRIS-CSE, and BioVidEmo perform poorly across all metrics, underscoring the challenges associated with video formats in FER tasks.

    The exploration and benchmarking of FER datasets allowed us to identify both strengths and weaknesses, from which we derived several recommendations for future FER research. For example, large-scale datasets (with at least 500 samples) such as AffectNet and FER2013 to ensure better generalization, while addressing class imbalances and ensuring demographic diversity to mitigate biases. Additionally, researchers are advised to avoid relying on video-sampled frames due to their inconsistent expression intensities and to perform Paired Similarity analyses when combining datasets to prevent redundancy. For new dataset construction, employing automatic data collection to capture extensive variability, incorporating diverse lighting, settings, and perspectives, including rich metadata, and eliminating duplicate images are crucial steps to enhance both model performance and fairness.
    
    Overall, this study contributes a robust framework for FER dataset evaluation and offers valuable guidelines that will inform future research aimed at developing more reliable and equitable facial expression recognition systems.

\section*{Acknowledgments}

    This work is part of the Project PID2023-149079OB-I00 (EXPLAINME) funded by MICIU/AEI/10.13039/ 501100011033/ and FEDER, EU and of Project PID2022-136779OB-C32 (PLEISAR) funded by MICIU/ AEI /10.13039/501100011033/ and FEDER, EU. F. X. Gaya-Morey was supported by an FPU scholarship from the Ministry of European Funds, University and Culture of the Government of the Balearic Islands.

    The authors used ChatGPT to improve the readability and language of the manuscript. They reviewed and edited the content as needed and take full responsibility for the content of the published article.

 
\bibliography{bibliography}
%

\bibliographystyle{IEEEtran}

\end{document}